\documentclass[make,article,submit,moreauthors]{Definitions/mdpi} 
\ifdefined\pdfobjcompresslevel
\fi

\ifdefined\pdfoutput
  \pdfoutput=1
\fi
\firstpage{1} 
\pubvolume{1}
\issuenum{1}
\articlenumber{0}
\pubyear{2026}
\copyrightyear{2026}
\datereceived{ } 
\daterevised{ } 
\dateaccepted{ } 
\datepublished{ } 

\Title{DySR-GNN: Dynamic Sensor-Relationship Graph Neural Network for Root Cause Analysis in Automotive HiL Multivariate Time Series}

\Author{Hamza Ouarrad $^{1,}$*\orcidA{}, Mohammad Abboush $^{1}$ and Andreas Rausch $^{1}$}

\AuthorNames{Hamza Ouarrad, Mohammad Abboush and Andreas Rausch}

\address{%
$^{1}$ \quad Institute for Software and Systems Engineering, 
Technische Universität Clausthal, 
38678 Clausthal-Zellerfeld, Germany; 
hamza.ouarrad@tu-clausthal.de (H.O.); 
mohammad.abboush@tu-clausthal.de (M.A.); 
andreas.rausch@tu-clausthal.de (A.R.)}

\corres{Correspondence: hamza.ouarrad@tu-clausthal.de}

\abstract{Root-cause analysis in automotive Hardware-in-the-Loop (HiL) systems is challenging because faults can propagate across multiple interacting signals, making the most abnormal sensor different from the actual fault origin. This work proposes DySR-GNN, a Dynamic Sensor-Relationship Graph Neural Network for supervised multi-label root-cause identification in multivariate HiL time series. The approach jointly extracts temporal patterns within individual sensor signals and structural relationships across sensors. Each sensor is processed by a shared temporal encoder and combined with a trainable sensor-identity representation, from which a sparse relationship graph is reconstructed independently for every input window. This window-specific structure provides an explicit representation of the sensor relationships used for graph message passing and direct root-location prediction. Under a recording-disjoint evaluation protocol, the proposed model outperforms CNN--LSTM, Static GCN, Static GAT, and context-based variants, achieving a recording-level Macro-F1 of 0.9524, MCC of 0.9302, and strict exact-match accuracy of 0.8519. Analysis of the learned graphs reveals both persistent and state-dependent relationships among sensor variables. The results demonstrate that extracting adaptive relational structure alongside temporal representations can improve multi-label root-cause identification in complex automotive cyber--physical systems.}


\begin{document}
\nolinenumbers


\section{Introduction}
\label{sec:introduction}

Modern vehicles are software-intensive cyber--physical systems in
which electronic control units, sensors, actuators, communication
interfaces, and software functions interact to realize increasingly
complex and safety-critical functionality. Their development therefore
requires rigorous verification and validation activities, particularly
for safety-related electrical and electronic systems governed by the
ISO~26262 functional-safety standard~\cite{iso26262part10}. Within this
development process, Hardware-in-the-Loop (HiL) \cite{hanselmann1993hil} testing provides a
controlled and repeatable environment in which implemented control
software and hardware can be exercised under realistic operating
conditions, including abnormal situations that would be difficult,
costly, or unsafe to reproduce directly in a physical vehicle.

A direct consequence of large-scale HiL validation is the generation
of high-dimensional multivariate time-series recordings. Measurements
from driver inputs, vehicle dynamics, powertrain components, control
variables, pressures, temperatures, and actuator states evolve
simultaneously and interact through the physical and software-controlled
system. As system complexity increases, manual inspection of these
recordings becomes increasingly difficult. More importantly, a
disturbance originating at one location can propagate through several
dependent variables before becoming visible at the system level.
Consequently, the signal exhibiting the strongest deviation is not
necessarily the signal at which the fault originated. This distinction
is central to \emph{root cause analysis} (RCA), whose objective is not
only to detect an abnormal state, but to identify the underlying source
responsible for the observed system behaviour~\cite{pietsch2024rca, rooney2004rca}.

Classical fault detection and diagnosis (FDD) approaches include
signal-based monitoring, predefined thresholds, expert rules, and
model-based residual analysis~\cite{isermann2005fdd}. Although these
approaches remain valuable in industrial validation, their performance
can depend strongly on manually engineered knowledge or sufficiently
accurate reference models. Data-driven methods provide an alternative
by learning diagnostic representations directly from historical
measurements. Machine learning (ML) and deep learning (DL) have
therefore become increasingly important for time-series analysis and
fault diagnosis, with recurrent and convolutional architectures
providing powerful mechanisms for learning temporal dependencies
\cite{hochreiter1997lstm, fawaz2019deepts}.

This temporal modeling capability has also dominated much of the
existing data-driven automotive FDD literature. Automotive HiL studies
have employed CNN--LSTM architectures for fault detection and
classification~\cite{abboush2022hil} and GRU-based representation
learning for unknown single and concurrent faults
\cite{abboush2023gru}. Such architectures are well suited to extracting
patterns along the temporal dimension and have demonstrated that
diagnostically relevant information can be learned directly from
sequential vehicle measurements. However, conventional sequence models
primarily represent the sensor channels within a Euclidean feature space
and do not explicitly expose the relationships through which information
is shared among interacting variables. Consequently, although
dependencies between sensors may be captured implicitly within the
learned representations, the model does not provide an explicit
relational structure describing which sensors are connected or most
relevant to one another for a given observation. In this sense, these
approaches primarily model \emph{how sensor signals evolve over time},
while the complementary question of \emph{how the sensors relate to one
another} remains only implicitly represented.

This distinction is particularly relevant in multivariate automotive
systems. For example, a perturbation of an accelerator-pedal signal may
subsequently influence throttle position, engine speed, torque,
vehicle speed, manifold pressure, and other quantities. Diagnostically
relevant information is therefore contained not only in the temporal
evolution of each signal but also in the structure of the dependencies
between signals. From a knowledge-extraction perspective, this problem
can be interpreted as the joint discovery of hidden temporal and
structural patterns in high-dimensional observations
\cite{holzinger2019make}. Recent research on graph neural networks for
time-series analysis similarly emphasizes the importance of modeling
both inter-temporal and inter-variable relationships
\cite{jin2024gnn4ts}.

Graphs provide a natural mathematical representation of the latter.
Graph Neural Networks (GNNs) extend neural representation learning to
graph-structured data~\cite{scarselli2009gnn}, while Graph
Convolutional Networks (GCNs) enable node representations to be updated
through neighborhood-based information aggregation
\cite{kipf2017gcn}. When applied to multivariate sensor data, each
signal can be represented as a graph node and the connections between
nodes can encode relationships through which information is exchanged.
This provides an explicit relational inductive bias that conventional
sequence models do not naturally expose. A major challenge, however, is that an appropriate sensor graph is
rarely known a priori. Physical connectivity alone may not fully
describe the relationships that are useful for a diagnostic task, and
a single static graph may be insufficient when the system operates
under changing conditions. Vehicle speed, driver commands, controller
responses, engine state, and fault manifestations vary throughout a
recording, potentially changing which inter-sensor relationships are
most informative. A growing body of GNN-based time-series research
therefore considers learned and adaptive graph structures rather than
assuming a fixed topology~\cite{jin2024gnn4ts,zheng2025dynamicgnn}.
GDN, for example, demonstrated that trainable sensor embeddings and
learned sensor relationships can improve multivariate anomaly detection
\cite{deng2021gdn}.

For RCA, the challenge is even more specific. A diagnostic model must
distinguish the variable at which a fault was directly introduced from
other variables that react to the resulting disturbance. Statistical
association alone is not sufficient to establish such a causal
interpretation. Causal graphical modeling explicitly distinguishes
association from cause--effect structure and requires assumptions
stronger than those required for predictive dependency learning
\cite{sucar2026causal}. In particular, probabilistic dependence between
two variables does not by itself determine the direction or existence
of a causal relation. Accordingly, the graph learned in this work is
not interpreted as a causal graph. Instead, it represents
task-relevant statistical sensor relationships that are optimized for
the supervised identification of known fault-injection locations.

The resulting research problem therefore contains two complementary
dimensions. First, the model must capture the
\emph{temporal dynamics within each sensor signal}. Second, it must
capture the \emph{relational structure between sensors} and allow that
structure to adapt to the current observation. This motivates an
architecture in which temporal representation learning and graph
structure learning are treated as distinct but jointly optimized
components.

To address this problem, this paper proposes
\emph{DySR-GNN}, a \emph{Dynamic Sensor-Relationship Graph Neural Network}
for supervised multi-label RCA of automotive HiL time series. Each sensor window is first processed by a shared one-dimensional convolutional temporal
encoder to extract a compact representation of its local dynamics.
The temporal representation is then combined with a trainable
sensor-identity embedding, allowing the model to distinguish
window-specific signal behaviour from persistent sensor identity.
The resulting node features are transformed into graph-state
representations from which pairwise sensor similarities are estimated.
Top-$k$ sparsification and symmetrization yield a sparse adjacency
matrix reconstructed independently for each input window.

The learned graph subsequently defines the neighborhood over which a
two-layer GCN performs message passing. Thus, the temporal encoder
models \emph{how individual sensor signals evolve}, whereas dynamic
graph construction determines \emph{which sensor representations
should interact for the current window}. Finally, the learned
representations of the predefined candidate root nodes are evaluated by
a shared scoring mechanism that produces independent root
probabilities. The multi-label formulation allows the same architecture
to represent both individual and concurrent fault locations.

The central hypothesis investigated in this work is that
\emph{jointly modeling temporal sensor behaviour and adaptive
inter-sensor structure improves root-source identification compared
with architectures that use temporal information alone or a fixed
sensor graph}. The contribution is therefore not a new graph
convolution operator and does not constitute causal discovery.
Instead, it lies in dynamically reconstructing a task-relevant
sensor-relationship graph from each multivariate observation and using
that representation for direct supervised root-location identification.

The principal contributions of this work are summarized as follows:

\begin{itemize}

    \item We formulate automotive HiL root cause analysis as a
    \textbf{supervised multi-label sensor-location problem}, supporting
    both individual and concurrent fault origins.

    \item We propose a \textbf{dual temporal--relational representation}
    in which a shared temporal encoder captures within-sensor dynamics,
    while graph structure learning explicitly captures inter-sensor
    relationships.

    \item We introduce a \textbf{window-specific dynamic
    sensor-relationship graph} constructed from temporal sensor
    representations and trainable sensor-identity embeddings, allowing
    the relational structure to adapt to the current multivariate
    observation.

    \item We combine the learned topology with
    \textbf{GCN-based message passing and shared root scoring},
    maintaining a clear separation between sensor-relationship learning
    and root-location prediction.

    \item We evaluate the proposed architecture using a
    \textbf{recording-disjoint experimental protocol} and compare it
    with temporal and graph-based baselines, including CNN--LSTM,
    Static GCN, and Static GAT models, while additionally assessing
    robustness across multiple random seeds.

    \item We analyze the \textbf{knowledge encoded by the learned graph}
    through window-specific topology and edge-persistence analysis and
    report computational complexity and inference latency to assess the
    practical cost of dynamic graph learning.

\end{itemize}

The remainder of this article is organized as follows.
Section~\ref{sec:related_work} reviews automotive HiL fault diagnosis,
graph-based multivariate time-series learning, and graph-based RCA.
Section~\ref{sec:methodology} presents the proposed methodology and its
mathematical formulation. The automotive HiL case study, data
preparation, and experimental protocol are subsequently described in Section~\ref{sec:case_study}.
The experimental results and learned graph behavior are then analyzed
and discussed in Section~\ref{sec:Results and Discussion}. Finally, the conclusions, limitations, and directions
for future research are presented.

\section{Related Work}
\label{sec:related_work}

The proposed approach lies at the intersection of three research areas:
data-driven fault diagnosis for automotive Hardware-in-the-Loop (HiL)
validation, graph-based learning for multivariate sensor time series,
and graph-based root cause analysis (RCA). This section reviews the
most relevant developments in these areas and clarifies the distinction
between existing fault-diagnosis and anomaly-detection approaches and
the sensor-level RCA formulation considered in this work.

\subsection{Data-Driven Fault Diagnosis in Automotive HiL Validation}
\label{subsec:related_hil}

Hardware-in-the-Loop simulation provides a controlled environment in
which faults can be injected during automotive software validation
without exposing a physical vehicle or test driver to hazardous
conditions \cite{abboush2022hardware}.At the same time, modern automotive validation generates large volumes
of multivariate sensor data, motivating increasingly data-driven
approaches to anomaly and fault diagnosis
\cite{fang2024anomaly}. Abboush et al.~\cite{abboush2022hil} investigated intelligent fault
detection and classification for automotive software validation using
real-time HiL fault injection. Their approach combined one-dimensional
convolutional neural networks and Long Short-Term Memory networks to
automatically extract temporal features from multivariate sensor
recordings and classify injected faults. This work demonstrated that
deep neural architectures can effectively automate the analysis of
automotive HiL recordings and established an important basis for
data-driven diagnosis in this application domain. Subsequent work addressed the more difficult case in which fault
classes are not necessarily known during model development.
Abboush et al.~\cite{abboush2023gru} combined a GRU-based denoising
autoencoder with clustering to detect and group unknown single and
concurrent faults during automotive system-integration testing.
The study demonstrated the feasibility of identifying simultaneous
fault manifestations from multivariate HiL recordings under noisy
conditions without requiring fully supervised fault labels.

More recently, explainable hybrid deep-learning architectures have been
investigated for fault detection, identification, and localization in
automotive software-system validation
\cite{abboush2026explainable}. Such approaches improve the ability to
associate complex multivariate observations with faulty components and
provide additional information about the signals contributing to a
diagnostic decision.

Although these studies demonstrate the effectiveness of deep learning
for automotive HiL analysis, the sensor channels are primarily treated
as components of a Euclidean multivariate input. Inter-sensor
relationships are therefore represented implicitly by the neural
network rather than explicitly through a learned graph structure.
This distinction is important for RCA, where abnormalities can propagate
from an injected root signal to multiple downstream variables. In such
situations, identifying abnormal signals alone does not necessarily
identify the original fault location.

\subsection{Graph Learning for Multivariate Sensor Time Series}
\label{subsec:related_graph}

Graph neural networks provide an alternative representation in which
individual sensor channels are modeled as nodes and their dependencies
as graph edges. This formulation is particularly attractive for
multivariate time series because message passing allows information
from related signals to be explicitly aggregated while preserving a
sensor-level representation. One influential approach is the Graph Deviation Network (GDN) proposed
by Deng and Hooi~\cite{deng2021gdn}. GDN associates each sensor with a
trainable embedding, learns relationships between sensors, and performs
graph-attention-based forecasting. Deviations between observed and
predicted behavior are subsequently used for anomaly detection and
diagnostic interpretation. In contrast to conventional sequence models,
GDN explicitly exposes learned inter-sensor relationships, while
Kovalenko et al.~\cite{kovalenko2024trainable} further investigated
trainable adjacency structures for fault diagnosis when sensor
dependencies are not known a priori.

Zhao et al.~\cite{zhao2020mtadgat} proposed MTAD-GAT, which employs
graph-attention mechanisms to model dependencies in both the temporal
and feature dimensions of multivariate time series. Forecasting and
reconstruction objectives are jointly optimized in a self-supervised
setting, enabling anomaly detection while capturing dependencies among
variables. Han and Woo~\cite{han2022fusagnet} subsequently introduced
a sparse latent graph representation for multivariate anomaly
detection, combining representation learning, graph structure learning,
and forecasting to exploit cross-variable dependencies.

These studies motivated further research into graph structures that
adapt to the observed data instead of remaining fixed. MSDG
\cite{zhao2024msdg}, for example, employs dynamic graph learning to
capture changing correlations in industrial multivariate time series.
Such methods recognize that the interaction structure among sensor
signals may vary with system state and that a single global adjacency
matrix can therefore be restrictive. Dynamic graph learning has also been increasingly adopted for explicit
fault-diagnosis tasks. Wang et al.~\cite{wang2023atsgnn} proposed an
attention-aware temporal--spatial GNN for multi-sensor fault diagnosis,
using graph learning to fuse structural and temporal information.
Duan et al.~\cite{duan2025dygat} introduced DyGAT-FTNet, where
time--frequency representations are used to dynamically construct
multi-sensor graphs that are subsequently processed by a dynamic graph
attention network. Other work
has explored causal intervention in learned sensor graphs to reduce
spurious dependencies in industrial fault diagnosis
\cite{liu2024causalintervention}. More recently, Gu et al.~\cite{gu2026dmsgnn}
proposed a Dynamic Multi-Scale Graph Neural Network for hydraulic-system
health monitoring, in which the sensor topology is inferred
adaptively rather than specified a priori.

Graph-based diagnosis is also receiving attention in the target journal
of the present study. Li et al.~\cite{li2025amh} proposed adaptive
multi-view hypergraph learning for cross-condition bearing fault
diagnosis. Their framework dynamically constructs high-order
relationships across multiple signal representations and demonstrates
the increasing role of adaptive relational modeling in intelligent
fault diagnosis.

Collectively, these works establish that graph learning can provide
substantial advantages over models that process sensor channels
independently or through fixed Euclidean structures. However, the
majority of existing methods formulate the downstream task as anomaly
detection or fault-state classification. Consequently, the learned
graph is generally optimized to distinguish normal from abnormal
behavior or to identify a fault category, rather than to directly
identify the sensor location at which a fault originated.

\subsection{Graph-Based Root Cause Analysis}
\label{subsec:related_rca}

Root cause analysis differs from conventional fault classification in
that the objective is not only to recognize that a system is operating
abnormally, but to identify the source that produced the observed
downstream effects. Graph representations are particularly attractive
for this purpose because they can encode dependencies and propagation
paths among system variables or components.

Graph-based RCA has been investigated across several interconnected
systems. Dong et al.~\cite{dong2023hierarchical} used hierarchical
causal graphs to identify fault origins and propagation paths in complex
industrial processes, while graph structure learning has also been
combined with GNNs for RCA in communication networks. In
\cite{yen2022gnnrca}, hidden dependencies among multivariate network
KPIs are inferred using graph structure learning and subsequently used
by a GNN to identify root-cause and affected nodes. This demonstrates
that learned relational structure can be directly exploited for
root-source identification rather than solely for anomaly detection.

More recently, ST-GraphRCA~\cite{su2026stgraphrca} employed dynamically
weighted propagation graphs for RCA in distributed IoT systems, while
Chen et al.~\cite{chen2026causalvehicle} introduced a causal-constraint
graph attention framework for anomaly interpretation and root-cause
diagnosis in intelligent vehicles. Both approaches illustrate the value
of relational structure for root-source localization, but their graph
semantics differ from those considered here. The former models
service-level propagation, while the latter incorporates causal
constraints; the present study instead learns window-specific
task-relevant statistical relationships directly from automotive HiL
time series.

The distinction between graph-based anomaly explanation and direct RCA
is also important. Methods such as GDN can rank variables associated
with an anomalous event, but such a ranking is derived from forecasting
deviations after an anomaly has been detected. In automotive HiL
validation, multiple downstream signals may react strongly to a
fault injected elsewhere in the system. A large deviation therefore
does not necessarily imply that the corresponding sensor is the
originating fault location.

\subsection{Positioning of the Present Work}
\label{subsec:related_positioning}

The reviewed literature reveals three complementary developments.
First, deep-learning methods have demonstrated that automotive HiL
recordings can support accurate data-driven fault detection,
classification, clustering, and localization. Second, graph-based
time-series methods have shown that explicitly modeling inter-sensor
dependencies can improve anomaly detection and fault diagnosis.
Third, recent RCA methods demonstrate that graph structure can support
root-source localization in complex interconnected systems.

The present work connects these directions through a supervised
sensor-level RCA formulation for automotive HiL time series. Each
sensor window is first encoded using a shared temporal encoder and
combined with a trainable sensor-identity embedding. Unlike approaches
that rely on a single fixed sensor topology, a sparse relationship
graph is reconstructed independently for each input window from the
current sensor representations. Standard GCN message passing is then
performed over this window-specific topology, while a shared root
scorer produces independent probabilities for the predefined candidate
fault locations.

Accordingly, the methodological contribution is not a new graph
convolution operator, nor is the learned graph interpreted as a causal
graph. Rather, the contribution lies in using a dynamically learned,
window-specific sensor relationship structure as an intermediate
representation for direct multi-label root cause identification.
This formulation differs from forecasting- and reconstruction-based
graph anomaly detectors, from graph-based fault-state classifiers, and
from topology-driven RCA methods. It is designed specifically to
distinguish directly injected root signals from correlated downstream
responses in multivariate automotive HiL recordings.

\section{Methodology}
\label{sec:methodology}

\begin{figure}[H]
\begin{adjustwidth}{-\extralength}{0cm}
    \centering
    \includegraphics[width=\fulllength]{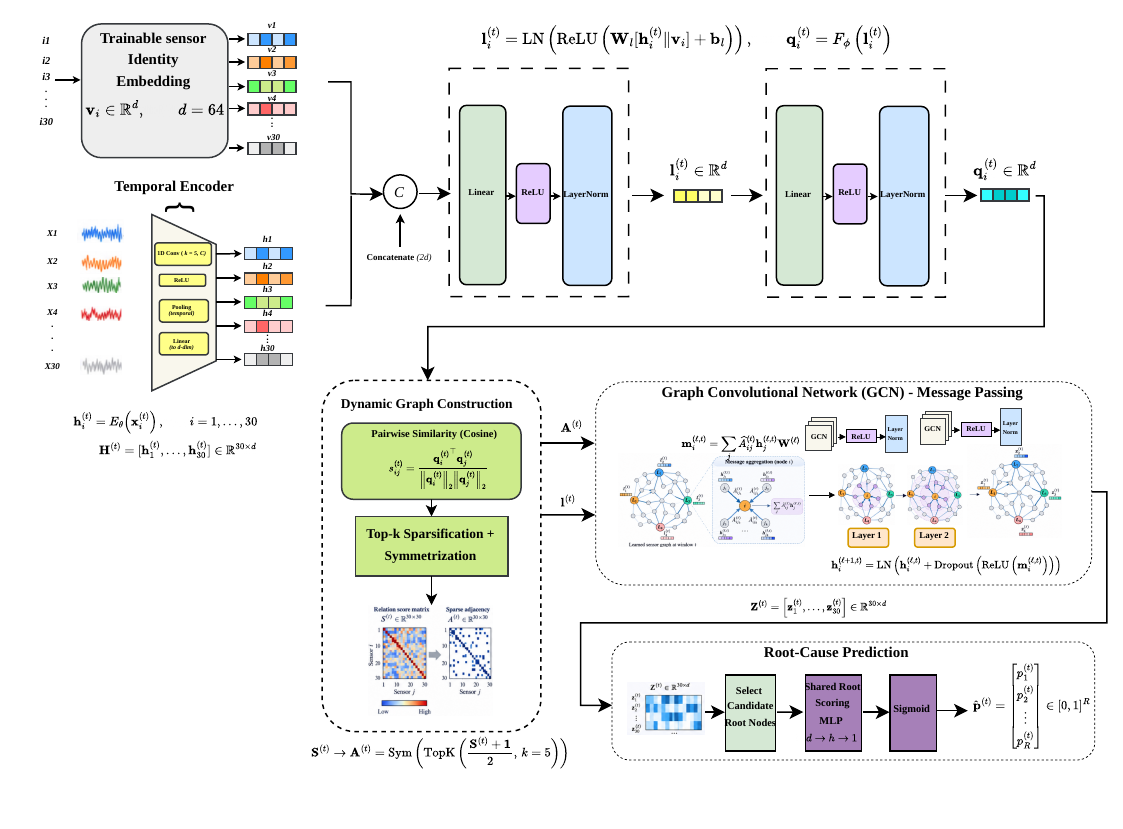}
    \caption{Architecture of the proposed DySR-GNN for multi-label
root-cause analysis of automotive HiL multivariate time series.
\label{fig:methodology}}
\end{adjustwidth}
\end{figure}

This section presents the problem formulation, data preparation, and the
proposed DySR-GNN. As shown in
Figure~\ref{fig:methodology}, the model combines sensor-wise temporal
encoding with trainable sensor identities, constructs a sparse
window-specific graph, and performs graph message passing for multi-label
root-cause prediction.

\subsection{Problem Formulation}
\label{subsec:problem_formulation}

The RCA task is formulated as supervised multi-label prediction over a
set of candidate root locations. Let a multivariate sensor window be
represented as

\begin{equation}
    \mathbf{X}^{(t)}
    =
    \left[
        \mathbf{x}_{1}^{(t)},
        \ldots,
        \mathbf{x}_{N}^{(t)}
    \right]
    \in
    \mathbb{R}^{N\times T},
\end{equation}

where $N$ denotes the number of monitored signals, $T$ the number of
temporal samples within a window, and $t$ the window index.

Rather than treating the $N$ signals only as a collection of feature
channels, they are represented as nodes of a graph. Following the
standard graph formulation $G=(V,E)$, where vertices denote objects and
edges encode relations between them~\cite{sucar2026causal}, the
sensor graph for window $t$ is written as

\begin{equation}
    \mathcal{G}^{(t)}
    =
    \left(
        \mathcal{V},
        \mathcal{E}^{(t)}
    \right),
\end{equation}

with

\begin{equation}
    \mathcal{V}
    =
    \left\{
        v_{1},\ldots,v_{N}
    \right\}.
\end{equation}

Each node $v_i$ corresponds to one monitored signal
$\mathbf{x}_{i}^{(t)}$, while $\mathcal{E}^{(t)}$ represents the
relationships used to exchange information between sensor nodes. The
node set remains fixed, whereas the edge structure is allowed to vary
with the observed window. The proposed architecture therefore seeks to
learn both a temporal representation of each signal and a
window-dependent relational representation of the multivariate system.

Given $R$ candidate root locations, the RCA target is defined as

\begin{equation}
    \mathbf{y}^{(t)}
    =
    \left[
        y_{1}^{(t)},
        \ldots,
        y_{R}^{(t)}
    \right]
    \in
    \{0,1\}^{R},
\end{equation}

where $y_{r}^{(t)}=1$ indicates that root location $r$ is active.
Multiple entries may be active simultaneously, allowing individual and
concurrent root causes to be represented within the same formulation.

The learning objective is to estimate

\begin{equation}
    f_{\Theta}:
    \mathbb{R}^{N\times T}
    \rightarrow
    [0,1]^{R},
\end{equation}

or, equivalently, for each candidate root,

\begin{equation}
    \hat{p}_{r}^{(t)}
    =
    P_{\Theta}
    \left(
        Y_{r}=1
        \mid
        \mathbf{X}^{(t)}
    \right),
    \qquad
    r=1,\ldots,R.
\end{equation}

Unlike approaches that first detect anomalous sensor behavior and infer
the root cause in a subsequent step, the proposed formulation learns
these root-location probabilities directly from the multivariate sensor
window. The problem consequently involves two complementary forms of
information: the temporal behavior within individual signals and the
relationships among signals. The following subsection describes how
both are learned jointly by the proposed dynamic graph architecture.

\subsection{Proposed DySR-GNN Architecture}
\label{subsec:dynamic_gnn}

The proposed architecture is designed around a simple premise: root-cause
information is contained not only in how individual sensor signals evolve,
but also in how their relationships change with the current system state.
Accordingly, the model separates temporal representation learning from
relationship learning. As illustrated in Figure~\ref{fig:methodology},
each sensor is first encoded independently, then embedded into a
window-dependent graph on which message passing is performed.

\subsubsection{Temporal and Sensor-Identity Representation}
\label{subsubsec:temporal_identity}

For sensor $i$ in window $t$, the temporal signal
$\mathbf{x}_{i}^{(t)} \in \mathbb{R}^{T}$ is processed by a shared
one-dimensional convolutional encoder $E_{\theta}$,

\begin{equation}
    \mathbf{h}_{i}^{(t)}
    =
    E_{\theta}
    \left(
        \mathbf{x}_{i}^{(t)}
    \right),
    \qquad
    \mathbf{h}_{i}^{(t)}
    \in
    \mathbb{R}^{d}.
\end{equation}

The same encoder is applied to every sensor. It consists of two
convolutional blocks: a $1\!\rightarrow\!32$ convolution with kernel
size $7$, followed by batch normalization, ReLU activation, and max
pooling; and a $32\!\rightarrow\!64$ convolution with kernel size $5$,
again followed by batch normalization and ReLU. Adaptive average pooling
removes the remaining temporal dimension, and a linear projection maps
the representation to $d=64$ dimensions.

Weight sharing is deliberate. The network learns a common mechanism for
extracting temporal patterns rather than maintaining a separate encoder
for every sensor. Yet temporal behavior alone does not identify which
physical signal produced a representation. Each sensor therefore receives
a trainable identity embedding

\begin{equation}
    \mathbf{v}_{i}
    \in
    \mathbb{R}^{d}.
\end{equation}

The temporal and identity representations play different roles:
$\mathbf{h}_{i}^{(t)}$ describes \emph{what the sensor is doing in the
current window}, whereas $\mathbf{v}_{i}$ encodes \emph{which sensor it
is}. Their concatenation is projected into a local node representation,

\begin{equation}
    \mathbf{l}_{i}^{(t)}
    =
    \mathrm{LN}
    \left(
        \mathrm{ReLU}
        \left(
            \mathbf{W}_{l}
            \left[
                \mathbf{h}_{i}^{(t)}
                \Vert
                \mathbf{v}_{i}
            \right]
            +
            \mathbf{b}_{l}
        \right)
    \right),
\end{equation}

where $\Vert$ denotes concatenation and $\mathrm{LN}(\cdot)$ denotes
layer normalization.

A second transformation maps the local representation into a graph-state
embedding,

\begin{equation}
    \mathbf{q}_{i}^{(t)}
    =
    F_{\phi}
    \left(
        \mathbf{l}_{i}^{(t)}
    \right)
    =
    \mathrm{LN}
    \left(
        \mathrm{ReLU}
        \left(
            \mathbf{W}_{q}\mathbf{l}_{i}^{(t)}
            +
            \mathbf{b}_{q}
        \right)
    \right).
\end{equation}

This separation is important. The local state
$\mathbf{l}_{i}^{(t)}$ is retained as the node feature used later for
message passing, whereas $\mathbf{q}_{i}^{(t)}$ is used specifically to
determine the graph structure.

\subsubsection{Dynamic Sensor-Relationship Graph Construction}
\label{subsubsec:dynamic_graph}

The graph is reconstructed independently for every input window. Rather
than assuming a predefined physical topology, pairwise relationships are
derived from the current graph-state embeddings. For sensors $i$ and
$j$, cosine similarity is computed as

\begin{equation}
    s_{ij}^{(t)}
    =
    \frac{
        \left(
            \mathbf{q}_{i}^{(t)}
        \right)^{\top}
        \mathbf{q}_{j}^{(t)}
    }{
        \left\|
            \mathbf{q}_{i}^{(t)}
        \right\|_{2}
        \left\|
            \mathbf{q}_{j}^{(t)}
        \right\|_{2}
        +
        \epsilon
    }.
\end{equation}

The similarities are mapped from $[-1,1]$ to non-negative edge weights,

\begin{equation}
    w_{ij}^{(t)}
    =
    \frac{s_{ij}^{(t)}+1}{2}.
\end{equation}

A dense similarity matrix would allow every sensor to exchange
information with every other sensor. Instead, each node retains only its
$k$ strongest neighbors, with $k=5$. The sparse directed graph is then
symmetrized using the maximum weight in either direction,

\begin{equation}
    A_{ij}^{(t)}
    =
    \max
    \left(
        \widetilde{A}_{ij}^{(t)},
        \widetilde{A}_{ji}^{(t)}
    \right).
\end{equation}

In compact form,

\begin{equation}
    \mathbf{A}^{(t)}
    =
    \operatorname{Sym}
    \left[
        \operatorname{TopK}
        \left(
            \frac{\mathbf{S}^{(t)}+1}{2},
            k=5
        \right)
    \right].
\end{equation}

Because the graph-state embeddings depend on the current temporal
window, the resulting topology is generally time-varying,

\begin{equation}
    \mathbf{A}^{(t)}
    \neq
    \mathbf{A}^{(t+1)}.
\end{equation}

Thus, the graph does not prescribe a permanent neighborhood for a sensor.
Some relationships may persist, while others are replaced as the
multivariate state changes. The learned adjacency represents
task-relevant statistical relationships and is not interpreted as a
causal graph.

\subsubsection{Graph Message Passing and Root Prediction}
\label{subsubsec:gcn_prediction}

Before message passing, self-loops are added and the adjacency matrix is
symmetrically normalized following the standard GCN formulation
\cite{kipf2017gcn},

\begin{equation}
    \widehat{\mathbf{A}}^{(t)}
    =
    \left(
        \mathbf{D}^{(t)}
    \right)^{-\frac{1}{2}}
    \left(
        \mathbf{A}^{(t)}
        +
        \mathbf{I}
    \right)
    \left(
        \mathbf{D}^{(t)}
    \right)^{-\frac{1}{2}}.
\end{equation}

The initial node-feature matrix contains the local sensor
representations,

\begin{equation}
    \mathbf{H}^{(0,t)}
    =
    \left[
        \mathbf{l}_{1}^{(t)},
        \ldots,
        \mathbf{l}_{N}^{(t)}
    \right]^{\top}.
\end{equation}

Two graph-convolutional layers are then applied. At layer $\ell$,

\begin{equation}
    \mathbf{M}^{(\ell,t)}
    =
    \widehat{\mathbf{A}}^{(t)}
    \mathbf{H}^{(\ell,t)}
    \mathbf{W}^{(\ell)}
    +
    \mathbf{b}^{(\ell)},
\end{equation}

followed by residual propagation,

\begin{equation}
    \mathbf{H}^{(\ell+1,t)}
    =
    \mathrm{LN}
    \left[
        \mathbf{H}^{(\ell,t)}
        +
        \mathrm{Dropout}
        \left(
            \mathrm{ReLU}
            \left(
                \mathbf{M}^{(\ell,t)}
            \right)
        \right)
    \right].
\end{equation}

The residual path preserves sensor-local information while the graph
branch incorporates information from neighboring nodes. After two
layers, the model obtains

\begin{equation}
    \mathbf{Z}^{(t)}
    =
    \mathbf{H}^{(2,t)}
    \in
    \mathbb{R}^{N\times d}.
\end{equation}

Root prediction is performed directly from the final representations of
the predefined candidate root nodes. For root $r$, the corresponding
node embedding $\mathbf{z}_{r}^{(t)}$ is passed through the same shared
scoring network,

\begin{equation}
    s_{r}^{(t)}
    =
    \mathbf{w}_{2}^{\top}
    \mathrm{Dropout}
    \left[
        \mathrm{ReLU}
        \left(
            \mathbf{W}_{1}
            \mathbf{z}_{r}^{(t)}
            +
            \mathbf{b}_{1}
        \right)
    \right]
    +
    b_{2},
\end{equation}

and converted into a root probability by

\begin{equation}
    p_{r}^{(t)}
    =
    \sigma
    \left(
        s_{r}^{(t)}
    \right).
\end{equation}

Applying the scorer independently to all $R$ candidate root nodes yields

\begin{equation}
    \widehat{\mathbf{p}}^{(t)}
    =
    \left[
        p_{1}^{(t)},
        \ldots,
        p_{R}^{(t)}
    \right].
\end{equation}

Independent sigmoid outputs are used instead of a Softmax layer because
the root locations are not mutually exclusive. The architecture can
therefore assign high probability to several roots within the same
window, which is required for concurrent-fault RCA.

\subsection{Training Objective}
\label{subsec:training_objective}

Because the candidate root locations are not mutually exclusive, the
network is trained as a multi-label classifier using independent sigmoid
outputs. Each root therefore contributes a separate binary prediction,
while several roots may be active within the same sample.

The model is optimized using a weighted binary cross-entropy objective,

\begin{equation}
    \mathcal{L}
    =
    -
    \frac{1}{BR}
    \sum_{b=1}^{B}
    \sum_{r=1}^{R}
    \left[
        \omega_r y_{br}\log(p_{br})
        +
        (1-y_{br})\log(1-p_{br})
    \right],
\end{equation}

where $B$ denotes the mini-batch size, $R$ the number of candidate root
locations, $y_{br}\in\{0,1\}$ the ground-truth label, and $p_{br}$ the
corresponding predicted root probability. The weight $\omega_r$
compensates for differences in the occurrence frequency of individual
root locations and is computed from the training partition only as

\begin{equation}
    \omega_r
    =
    \frac{N_r^{-}}
    {\max(N_r^{+},1)},
\end{equation}

with $N_r^{+}$ and $N_r^{-}$ denoting the number of positive and
negative training samples for root $r$, respectively.

This formulation penalizes each candidate root independently and thus
preserves the multi-label nature of the RCA task. Model selection is
based on the validation loss, with the checkpoint achieving the lowest
validation loss retained for evaluation. The same optimization protocol
is applied to the proposed model and all reference and ablation models
to maintain a consistent comparison.

The principal training settings used throughout the experiments are
summarized in Table~\ref{tab:training_settings}.

\begin{table}[H]
\centering
\caption{Training and implementation settings used in the experiments.}
\label{tab:training_settings}
\begin{tabular}{ll}
\hline
\textbf{Setting} & \textbf{Value} \\
\hline
Optimizer & AdamW \\
Initial learning rate & $1\times10^{-3}$ \\
Weight decay & $1\times10^{-4}$ \\
Batch size & 32 \\
Maximum epochs & 150 \\
Early-stopping patience & 20 epochs \\
Learning-rate scheduler & ReduceLROnPlateau \\
Scheduler reduction factor & 0.5 \\
Scheduler patience & 5 epochs \\
Gradient clipping & Maximum norm 5.0 \\
Embedding dimension $d$ & 64 \\
GCN hidden dimension & 64 \\
Root-scoring hidden dimension & 32 \\
Dropout probability & 0.30 \\
Dynamic graph neighborhood $k$ & 5 \\
Decision threshold $\tau$ & 0.50 \\
Random seeds & 42, 123, 2026 \\
\hline
\end{tabular}
\end{table}

\section{Case Study and Experimental Setup}
\label{sec:case_study}

\subsection{HiL Platform, Fault Injection, and Dataset}
\label{subsec:hil_dataset}

The experimental study is based on an automotive Hardware-in-the-Loop
(HiL) environment in which the behavior of the vehicle and its
subsystems is executed in real time while the control system interacts
with the simulated plant in closed loop. HiL testing provides a
repeatable laboratory setting in which operating conditions can be
reproduced and abnormal system behavior can be investigated without
requiring physical road tests. The experimental platform employs the
dSPACE SCALEXIO real-time simulation system, which is designed for
HiL testing of automotive electronic control systems
\cite{dspace_scalexio}.

Faulty operating conditions were generated using the real-time
fault-injection concept developed by Abboush et al.
\cite{abboush2022hardware}. The framework integrates fault injection
with the HiL execution environment, allowing selected system variables
to be manipulated during runtime without restructuring the underlying
vehicle model. This makes it possible to observe how a locally injected
disturbance is reflected across multiple interacting signals under
controlled and repeatable conditions. Such a setting is particularly
relevant for root-cause analysis: the injected location is known,
whereas its effects may propagate through several measured variables.

Four candidate root locations are considered in the present study,
covering driver-input, vehicle-control, and powertrain-related signals.
They are denoted by $L1$--$L4$:

\begin{itemize}
    \item $L1$: accelerator-pedal position
          (\texttt{Pos\_AccPedal});
    \item $L2$: steering-wheel angle
          (\texttt{Angle\_SteeringWheel});
    \item $L3$: engine speed
          (\texttt{n\_Engine});
    \item $L4$: throttle position
          (\texttt{Pos\_Throttle}).
\end{itemize}

The dataset is a multivariate time series data that contains one healthy reference recording and nine faulty
recordings. Each recording spans approximately $0$--$100\,\mathrm{s}$
and contains about $96{,}000$ observations, corresponding to an
effective acquisition rate close to $1\,\mathrm{kHz}$. For the faulty
recordings, the fault injection is active between $25$ and
$65\,\mathrm{s}$, while the intervals before and after the injection
represent non-faulted system operation within the same experiment.

Importantly, the faulty cases are not restricted to isolated roots.
Several recordings contain two simultaneously active root locations,
providing the multi-label conditions required to evaluate RCA under
concurrent faults. The available scenarios are summarized in
Table~\ref{tab:fault_scenarios}.

\begin{table}[H]
\centering
\caption{HiL recordings and corresponding active root locations.}
\label{tab:fault_scenarios}
\begin{tabular}{ll}
\hline
\textbf{Recording} & \textbf{Active root location(s)} \\
\hline
H      & Healthy reference \\
L1     & $L1$ \\
L2     & $L2$ \\
L3     & $L3$ \\
L1L2   & $L1$, $L2$ \\
L1L3   & $L1$, $L3$ \\
L1L4   & $L1$, $L4$ \\
L2L3   & $L2$, $L3$ \\
L2L4   & $L2$, $L4$ \\
L3L4   & $L3$, $L4$ \\
\hline
\end{tabular}
\end{table}

The supervised RCA experiments are based
on the nine faulty recordings, for which the injected root locations
provide the ground-truth multi-label targets. The healthy recording is
retained as a reference for the underlying system behavior. The raw HiL
recordings contain a considerably larger set of measured and commanded
signals; their selection and transformation into fixed-length model
inputs are described in the following subsection.

\subsection{Signal Selection, Data Preprocessing, and Window Generation}
\label{subsec:data_preprocessing}

The raw HiL recordings contain approximately 60 measured and commanded
signals. Not all of them provide useful information for relational
modeling. Globally constant channels and strongly redundant variables
were therefore removed, yielding a common set of $N=30$ signals. Each
retained signal corresponds to one graph node and is kept consistently
across all recordings.

The original measurements were acquired at approximately
$1\,\mathrm{kHz}$. To reduce computational load while preserving the
dynamics relevant to the considered fault scenarios, the signals were
resampled to $100\,\mathrm{Hz}$. Finite-impulse-response filtering was
applied before decimation to suppress frequency components above the
new Nyquist limit and avoid aliasing. Supervised samples were extracted only from the interval in which the
fault injection was active, i.e., from $25$ to $65\,\mathrm{s}$. Rather
than using isolated time points, the multivariate recordings were
segmented using overlapping sliding windows of $2\,\mathrm{s}$ with a
stride of $1\,\mathrm{s}$. At the resulting sampling rate, each window
contains

\begin{equation}
    T
    =
    2\,\mathrm{s}
    \times
    100\,\mathrm{Hz}
    =
    200
\end{equation}

samples per sensor. Only windows fully contained within the fault-active
interval were retained. This produces 39 windows for each faulty
recording and, across the nine faulty recordings, a total of 351
multivariate samples,

\begin{equation}
    \mathbf{X}^{(t)}
    \in
    \mathbb{R}^{30\times200}.
\end{equation}

The overlap between consecutive windows preserves temporal continuity
while exposing the model to different local realizations of the same
fault episode. Importantly, these windows are not treated as independent
experiments during data partitioning; all windows originating from one
HiL recording remain assigned to the same training, validation, or test
partition, as described in the following subsection.

Sensor-wise normalization is performed using statistics estimated from
the training recordings only. For sensor $i$,

\begin{equation}
    \widetilde{x}_{i}
    =
    \frac{
        x_i-\mu_i^{\mathrm{train}}
    }{
        \sigma_i^{\mathrm{train}}+\epsilon
    },
\end{equation}

where $\mu_i^{\mathrm{train}}$ and
$\sigma_i^{\mathrm{train}}$ denote the corresponding training-set mean
and standard deviation. The same parameters are subsequently applied to
the validation and held-out test recordings, preventing information
from these recordings from influencing the preprocessing stage.

\subsection{Experimental Protocol, Reference Models, and Evaluation}
\label{subsec:experimental_protocol}

The evaluation was designed at the recording level rather than at the
window level. This distinction is essential because consecutive windows
from the same HiL run overlap in time and share the same underlying
fault episode. Randomly distributing such windows across training and
test sets would therefore expose the model to closely related samples
from the same experiment and could lead to overly optimistic estimates
of generalization.

A fixed recording-disjoint protocol was consequently adopted. In each
fold, seven faulty recordings are used for training, one complete
recording for validation, and one complete recording for testing. All 39
windows generated from a given recording remain in the same partition.
Across nine folds, every faulty recording serves exactly once as the
held-out test case. The validation recording is changed accordingly, as
summarized in Table~\ref{tab:recording_folds}.

\begin{table}[H]
\centering
\caption{Recording-disjoint training, validation, and test protocol.}
\label{tab:recording_folds}
\begin{tabular}{ccc}
\hline
\textbf{Fold} & \textbf{Validation recording} & \textbf{Test recording} \\
\hline
1 & L2   & L1   \\
2 & L3   & L2   \\
3 & L1   & L3   \\
4 & L3L4 & L1L2 \\
5 & L2L4 & L1L3 \\
6 & L2L3 & L1L4 \\
7 & L1L4 & L2L3 \\
8 & L1L3 & L2L4 \\
9 & L1L2 & L3L4 \\
\hline
\end{tabular}
\end{table}

The remaining seven recordings in each fold constitute the training
partition. To account for stochastic variation during optimization,
each fold is repeated with three random seeds: 42, 123, and 2026.
Model selection relies exclusively on the corresponding validation
recording.

DySR-GNN is compared against models
chosen to separate the contributions of temporal modeling, graph-based
reasoning, and adaptive topology. A CNN--LSTM serves as a non-graph
temporal baseline.

The Static GCN follows the same general sensor representation pipeline
as the proposed model, including the shared temporal encoder,
trainable sensor-identity embeddings, local feature fusion, and shared
root-scoring mechanism. Its sensor graph, however, is constructed only
from the trainable identity embeddings. Pairwise cosine similarities are
computed between sensor embeddings, followed by the same top-$k$
sparsification with $k=5$ and graph symmetrization used in the proposed
model. The resulting sparse adjacency is shared by all input windows and
therefore represents a fixed learned sensor topology. The Static GAT retains this same static sparse graph and sensor
representation pipeline but replaces conventional graph convolution
with graph-attention-based neighborhood aggregation
\cite{velickovic2018graph}. It therefore evaluates whether a more
expressive aggregation mechanism can account for the observed
performance differences without adapting the topology itself. In
contrast, the proposed model reconstructs the sensor graph from the
current temporal and sensor-identity representations independently for
every input window.

Two additional variants assess whether explicit global context provides
further benefit. The first augments the dynamic architecture with a
dedicated context representation; the second introduces a gated context
fusion mechanism. These variants are evaluated under the same data
partitions and optimization conditions as the main model. The resulting
comparison therefore addresses three distinct questions: whether graph
reasoning improves over purely temporal modeling, whether adapting the
topology is beneficial relative to a static graph, and whether an
additional explicit context pathway improves upon the dynamic graph
representation alone.

Predictions are first produced at window level. For a held-out recording
containing $M$ windows, the root probabilities are subsequently
aggregated by averaging,

\begin{equation}
    \bar{p}_{r}
    =
    \frac{1}{M}
    \sum_{t=1}^{M}
    p_{r}^{(t)},
\end{equation}

where $p_{r}^{(t)}$ denotes the predicted probability for candidate
root $r$ in window $t$. In the present setup, $M=39$. A root location
is assigned to the recording when

\begin{equation}
    \bar{p}_{r} \geq \tau,
    \qquad
    \tau = 0.5.
\end{equation}

Performance is evaluated at both window and recording level. Macro-F1
and the Matthews correlation coefficient (MCC) quantify label-wise
classification quality, while strict exact-match accuracy measures
whether the complete multi-label root set is recovered correctly.
Exact match is particularly demanding in concurrent-fault cases: a
prediction is counted as correct only when every active and inactive
root label is assigned correctly. The Brier score is additionally
reported to quantify the squared error of the predicted root
probabilities.

Because windows from the same HiL experiment are not statistically
independent, inferential comparisons are performed at the recording
level. Predictions are first averaged across the three seeds for each
held-out recording. Pairwise differences between the proposed dynamic
model and the static graph baselines are then evaluated using an exact
paired sign-flip permutation test. In parallel, 95\% confidence
intervals for the mean paired differences are estimated by bootstrap
resampling the independent test recordings.

The statistical analysis is intentionally based on the nine held-out
recordings rather than the 351 overlapping windows. This yields a much
more conservative assessment of model differences, although the small
number of independent experiments necessarily limits statistical power.
The statistical results are therefore interpreted together with the
effect direction and descriptive performance rather than from
$p$-values alone.

\section{Results and Discussion}
\label{sec:Results and Discussion}

\subsection{Overall Root Cause Identification Performance}
\label{subsec:overall_results}

Table~\ref{tab:overall_results} summarizes the performance of the investigated models at window and recording level. Since neighboring windows originate from the same HiL experiment and are strongly correlated, the recording-level results provide a more relevant measure of whether a model can correctly identify the root causes of an unseen diagnostic case.

\begin{table}[H]
\caption{Performance comparison across three independent random seeds.
Results are reported as mean $\pm$ standard deviation.
\label{tab:overall_results}}

\begin{adjustwidth}{-\extralength}{0cm}
\begin{tabularx}{\fulllength}{lCCCCCC}
\toprule

\textbf{Model} &
\shortstack{\textbf{Window}\\\textbf{Macro-F1}} &
\shortstack{\textbf{Window}\\\textbf{MCC}} &
\shortstack{\textbf{Recording}\\\textbf{Macro-F1}} &
\shortstack{\textbf{Recording}\\\textbf{MCC}} &
\shortstack{\textbf{Exact}\\\textbf{Match}} &
\shortstack{\textbf{Brier}\\\textbf{Score}} \\

\midrule

CNN--LSTM
& $0.5116 \pm 0.0189$
& $0.1308 \pm 0.0501$
& $0.4888 \pm 0.0561$
& $0.1445 \pm 0.0928$
& $0.0741 \pm 0.0642$
& $0.2684 \pm 0.0053$ \\

Static GCN
& $0.8776 \pm 0.0372$
& $0.7910 \pm 0.0619$
& $0.8911 \pm 0.0300$
& $0.8225 \pm 0.0410$
& $0.6667 \pm 0.0000$
& $0.0686 \pm 0.0089$ \\

Static GAT
& $0.8830 \pm 0.0131$
& $0.7960 \pm 0.0239$
& $0.9157 \pm 0.0304$
& $0.8560 \pm 0.0615$
& $0.7407 \pm 0.0642$
& $0.0706 \pm 0.0185$ \\

\textbf{DySR-GNN}
& $\boldsymbol{0.8940 \pm 0.0057}$
& $\boldsymbol{0.8211 \pm 0.0079}$
& $\boldsymbol{0.9524 \pm 0.0206}$
& $\boldsymbol{0.9302 \pm 0.0302}$
& $\boldsymbol{0.8519 \pm 0.0642}$
& $\boldsymbol{0.0605 \pm 0.0034}$ \\

Context-aware variant
& $0.8764 \pm 0.0209$
& $0.7782 \pm 0.0396$
& $0.9098 \pm 0.0214$
& $0.8523 \pm 0.0449$
& $0.7407 \pm 0.0642$
& $0.0687 \pm 0.0160$ \\

Gated context-aware variant
& $0.8853 \pm 0.0175$
& $0.8046 \pm 0.0286$
& $0.9223 \pm 0.0182$
& $0.8760 \pm 0.0354$
& $0.7778 \pm 0.1111$
& $0.0687 \pm 0.0135$ \\

\bottomrule
\end{tabularx}
\end{adjustwidth}
\end{table}

A clear separation is observed between the temporal-only baseline and the graph-based models. The CNN-LSTM reaches a recording-level Macro-F1 of $0.4888\pm0.0561$, whereas the Static GCN increases this value to $0.8911\pm0.0300$. The corresponding MCC increases from $0.1445$ to $0.8225$. This large difference suggests that, for the present RCA task, learning temporal patterns alone is not sufficient. Fault effects are distributed across multiple interacting signals, and representing these signals as an explicitly connected system provides a considerably stronger diagnostic representation.

The improvement obtained by the Static GAT over the Static GCN is smaller but still consistent. Its recording-level Macro-F1 increases to $0.9157\pm0.0304$ and MCC to $0.8560\pm0.0615$. This indicates that the way information is aggregated from neighboring sensors also matters. Nevertheless, DySR-GNN achieves a further increase to $0.9524\pm0.0206$ Macro-F1 and $0.9302\pm0.0302$ MCC despite using a standard graph-convolutional operator. This pattern is important because it suggests that the gain cannot be attributed solely to a more expressive neighborhood aggregation mechanism; adapting the sensor topology itself appears to provide additional useful information for the RCA objective.

The same tendency becomes more pronounced under the strict exact-match criterion. Exact match requires the complete predicted root set to correspond to the ground truth and therefore penalizes both missing and spurious root locations. This is particularly relevant for the present multi-label setting, where concurrent faults must be identified jointly rather than independently. The CNN--LSTM correctly recovers the complete root set in only $7.4\%$ of cases on average, compared with $66.7\%$ for the Static GCN and $74.1\%$ for the Static GAT. The proposed model increases this value to $85.2\%$. Thus, the advantage of the dynamic formulation is not limited to improving individual root-label scores; it is also reflected in a substantially higher rate of correctly recovering the complete fault configuration.

\begin{figure}[H]
    \centering
    \includegraphics[width=\textwidth]
    {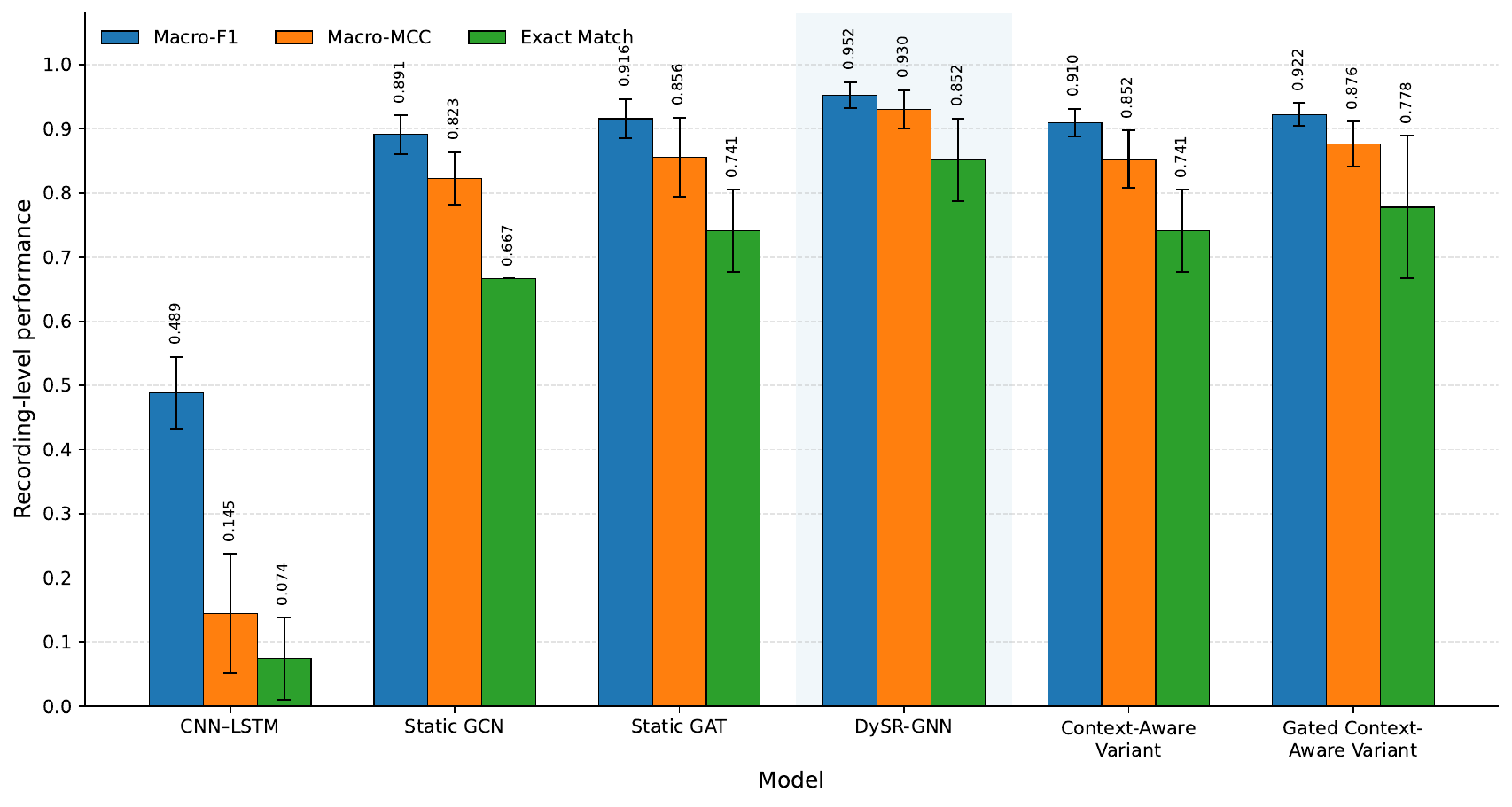}
    \caption{Recording-level root cause identification performance
    across three independent random seeds. Bars show the mean
    Macro-F1, MCC, and strict exact-match accuracy, while error
    bars indicate one standard deviation.
    \label{fig:overall_performance}}
\end{figure}

Figure~\ref{fig:overall_performance} further shows that the performance
ordering is consistent across the three principal recording-level
metrics. In particular, DySR-GNN improves Macro-F1 by
approximately $6.1$ percentage points over the Static GCN and
$3.7$ percentage points over the Static GAT. For exact-match accuracy,
the corresponding gains are approximately $18.5$ and $11.1$
percentage points, respectively. The larger differences observed for
exact match indicate that dynamic relationship modeling is especially
useful when the model must recover the complete combination of active
root locations. The probability estimates follow the same general pattern. The proposed
model obtains the lowest Brier score ($0.0605\pm0.0034$), compared with $0.0686\pm0.0089$ for the Static
GCN and $0.0706\pm0.0185$ for the Static GAT. Although these
differences are comparatively small and should not by themselves be
interpreted as evidence of superior calibration, they indicate that the
higher classification performance is not accompanied by a degradation
in probabilistic prediction quality.

Finally, the context-aware and gated variants do not improve upon the
simpler dynamic formulation. This observation suggests that the benefit
does not arise merely from increasing architectural complexity. Instead,
the results point toward the window-specific sensor relationship
structure as the principal modeling component associated with the
observed performance gain. The contribution of this mechanism and the
effect of the investigated architectural variants are examined in more
detail in the following subsection.

\subsection{Effect of Dynamic Sensor-Relationship Modeling}
\label{subsec:dynamic_effect}

The comparison between the static and dynamic graph models provides a direct view of the contribution of adaptive sensor relationships. Both models use sensor-wise temporal representations and sensor identity information. The difference lies in the topology: the Static GCN applies one learned graph to all inputs, whereas the proposed model reconstructs $\mathbf{A}^{(t)}$ for every window. This change has a clear effect at the recording level. Macro-F1 increases from $0.8911$ to $0.9524$, while exact-match accuracy rises from $0.6667$ to $0.8519$. The proposed model also remains ahead of the Static GAT. Hence, a more expressive aggregation rule alone does not explain the observed gain. Allowing the neighborhood structure itself to change with the current temporal state appears to be the more important difference. Such behavior is plausible in HiL recordings. Relationships among driver commands, actuator responses, vehicle dynamics, and powertrain variables are not equally informative at every moment. A fixed graph must represent all operating states with the same set of neighbors. The dynamic model, in contrast, can retain recurring relationships while replacing others as the sensor state changes.

The ablation results reinforce this interpretation. The explicit context-aware variant achieved a recording-level Macro-F1 of $0.9098\pm0.0214$, and the gated variant reached $0.9223\pm0.0182$, both below the proposed model
($0.9524\pm0.0206$). In the present setting, the temporal sensor representations already seem to carry much of the information needed to characterize the operating state. Likewise, the residual GCN provides a
sufficient balance between local and relational information without an additional gating mechanism.

Overall, the strongest configuration is also the simpler one. Its main advantage comes from reconstructing the sensor topology from the current temporal state rather than from adding further context or fusion components. The learned adjacency is therefore treated as a task-specific predictive relationship structure and not as a physical or causal graph.

\subsection{Robustness and Statistical Analysis}
\label{subsec:statistical_results}

The results were repeated across three independent random seeds to assess the stability of the observed performance. DySR-GNN showed comparatively low variation, reaching $0.9524\pm0.0206$ Macro-F1 and $0.9302\pm0.0302$ MCC at the recording level. Its Brier score was also the lowest among the evaluated models ($0.0605\pm0.0034$), compared with $0.0686\pm0.0089$ for the Static GCN and $0.0706\pm0.0185$ for the Static GAT. The difference in Brier score is modest; importantly, however, the improvement in classification performance is not accompanied by less reliable probability estimates.

Statistical testing was performed on the nine held-out recordings rather than on individual windows, since overlapping windows from the same HiL experiment are not independent observations. For each recording, predictions were first averaged across the three seeds and then compared
pairwise between models.

For the paired analysis, an F1 score was additionally computed for each
individual recording across its complete set of root labels. For
recording $j$, this recording-wise F1 is defined as

\begin{equation}
F1_{\mathrm{rec},j}
=
\frac{2TP_j}
{2TP_j+FP_j+FN_j},
\end{equation}

where $TP_j$, $FP_j$, and $FN_j$ are determined across the four
candidate root locations of that recording. This example-based measure
provides one scalar F1 value per independent recording and is therefore
used for the paired statistical comparisons. It is distinct from the
recording-level Macro-F1 reported in the overall performance analysis,
which is calculated label-wise across all held-out recordings and then
macro-averaged over the root locations.

\begin{table}[H]
\caption{Paired recording-level comparison between DySR-GNN and the static graph baselines. Confidence intervals were obtained
by bootstrap resampling of recordings.
\label{tab:statistical_results}}

\begin{adjustwidth}{-\extralength}{0cm}
\begin{tabularx}{\fulllength}{llCCCC}
\toprule

\textbf{Comparison} &
\textbf{Metric} &
\shortstack{\textbf{Mean}\\\textbf{Difference}} &
\shortstack{\textbf{95\% CI}\\\textbf{Low}} &
\shortstack{\textbf{95\% CI}\\\textbf{High}} &
\textbf{$p$-value} \\

\midrule

DySR-GNN vs. Static GCN
& Recording-wise F1
& $+0.0963$
& $-0.0123$
& $0.2543$
& $0.2500$ \\

DySR-GNN vs. Static GCN
& Exact Match
& $+0.1852$
& $-0.0000$
& $0.4074$
& $0.2500$ \\

DySR-GNN vs. Static GCN
& Brier Score
& $-0.0081$
& $-0.0188$
& $0.0028$
& $0.2031$ \\

\midrule

DySR-GNN vs. Static GAT
& Recording-wise F1
& $+0.0321$
& $0.0000$
& $0.0642$
& $0.2500$ \\

DySR-GNN vs. Static GAT
& Exact Match
& $+0.1111$
& $0.0000$
& $0.2222$
& $0.2500$ \\

DySR-GNN vs. Static GAT
& Brier Score
& $-0.0101$
& $-0.0285$
& $0.0040$
& $0.3789$ \\

\bottomrule
\end{tabularx}
\end{adjustwidth}
\end{table}

As shown in Table~\ref{tab:statistical_results}, The paired comparisons consistently favor the dynamic formulation at the descriptive level. Relative to the Static GCN, exact-match accuracy improves by approximately $18.5$ percentage points; compared with the
Static GAT, the gain is approximately $11.1$ percentage points. However, none of the paired tests reaches conventional statistical significance.

The limited number of independent test recordings is important here. With only nine paired cases, even relatively large performance differences are difficult to establish statistically. The results should therefore be interpreted as a consistent empirical advantage of dynamic graph construction rather than as evidence of statistically significant superiority. A larger collection of independently acquired HiL recordings would be required for a more conclusive statistical assessment.

\subsection{Sensitivity to Graph Sparsity}
\label{subsec:k_sensitivity}

\begin{figure}[H]
    \centering
    \includegraphics[width=0.78\textwidth]
    {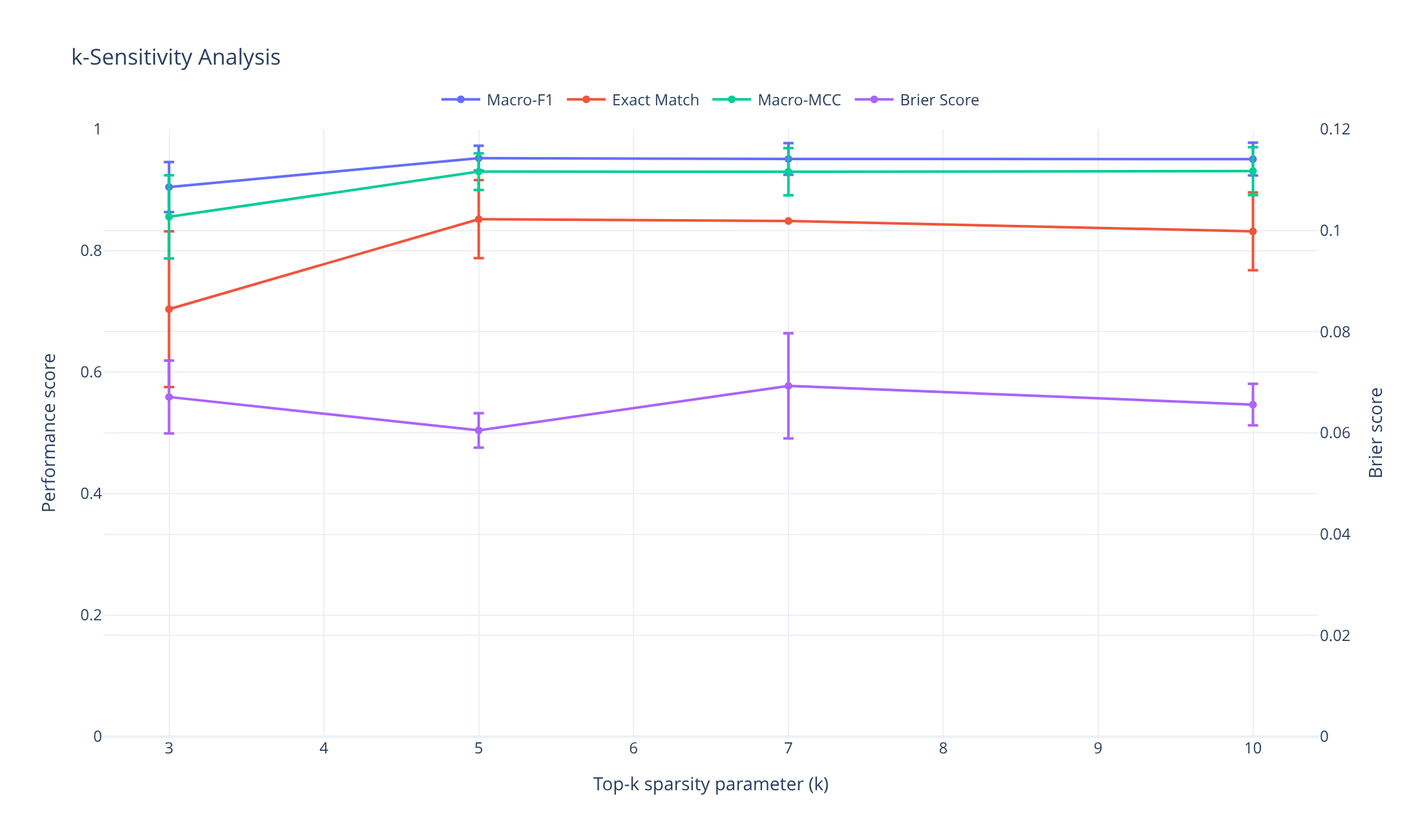}
    \caption{Sensitivity of DySR-GNN to the number of
    retained neighbors $k$. Results are reported at recording level
    across three random seeds.}
    \label{fig:k_sensitivity}
\end{figure}

To assess the robustness of the dynamic graph construction to the
sparsification parameter, the number of retained neighbors was varied
over $k\in\{3,5,7,10\}$ while keeping all remaining experimental
settings unchanged. Figure~\ref{fig:k_sensitivity} summarizes the
recording-level performance across three random seeds.

Performance decreases for $k=3$, with recording Macro-F1 falling to
$0.9048\pm0.0412$, indicating that excessive sparsification can remove
useful sensor relationships. In contrast, $k=5$, $k=7$, and $k=10$
produce closely comparable results, with Macro-F1 remaining around
$0.95$. Exact-match accuracy follows the same general trend. These
results indicate that the proposed method is not strongly dependent on
a narrowly tuned graph density once a sufficient number of neighbors is
retained. The choice $k=5$ is therefore maintained as a comparatively
sparse configuration without a meaningful loss in predictive
performance.

\subsection{Analysis of Learned Sensor Relationships}
\label{subsec:learned_graphs}

\begin{figure}[H]
    \centering
    \includegraphics[width=\textwidth]
    {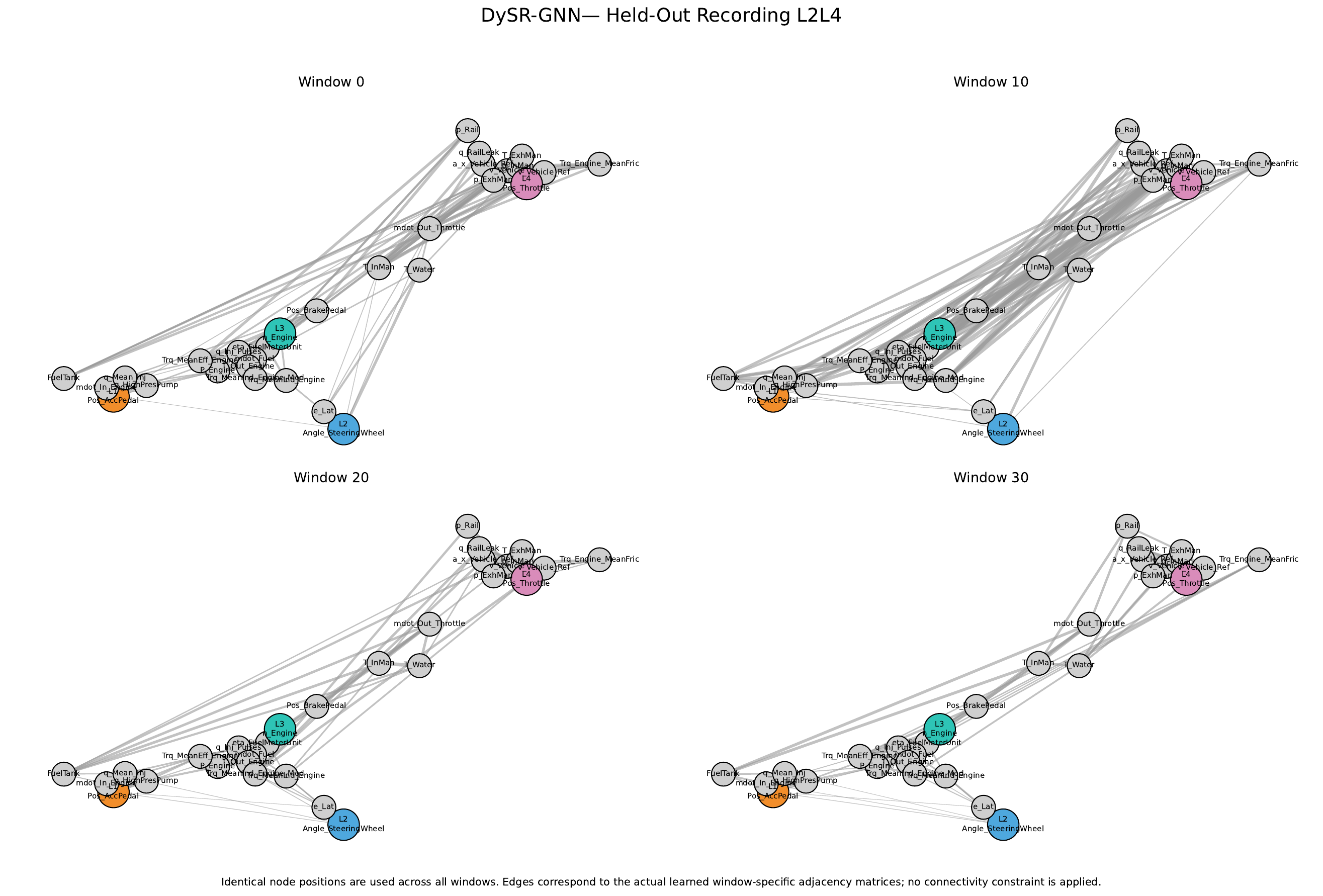}
    \caption{
        Four window-specific sensor-relationship graphs learned for the
        same held-out $L2L4$ recording. Identical node positions are
        used across panels to highlight changes in the learned topology.
        \label{fig:dynamic_graph_examples}
    }
\end{figure}

Beyond predictive performance, the dynamic adjacency matrices provide a direct view of the sensor relationships used by the model during inference. Unlike the static baselines, the proposed architecture produces a separate graph $\mathbf{A}^{(t)}$ for every input window.

Figure~\ref{fig:dynamic_graph_examples} shows that the learned topology changes across windows. Some edges remain visible over several temporal states, whereas others appear or disappear as the sensor representations evolve. The graph is therefore neither fully fixed nor reconstructed without continuity; rather, it combines recurring
relationships with state-dependent changes.

To quantify the persistence of learned sensor relationships,
we define the edge-selection frequency $f_{ij}$ as

\begin{equation}
    f_{ij}
    =
    \frac{1}{M}
    \sum_{t=1}^{M}
    \mathbb{I}
    \left[
        A_{ij}^{(t)} > 0
    \right],
    \label{eq:edge_frequency}
\end{equation}
where $M$ denotes the total number of windows,
$A_{ij}^{(t)}$ is the adjacency value between sensors
$i$ and $j$ in window $t$, and $\mathbb{I}[\cdot]$
is the indicator function. A value close to one
corresponds to a relationship selected repeatedly, while intermediate
values identify edges that are active only during parts of the
recording.

\begin{figure}[H]
    \centering
    \includegraphics[width=0.95\textwidth]
    {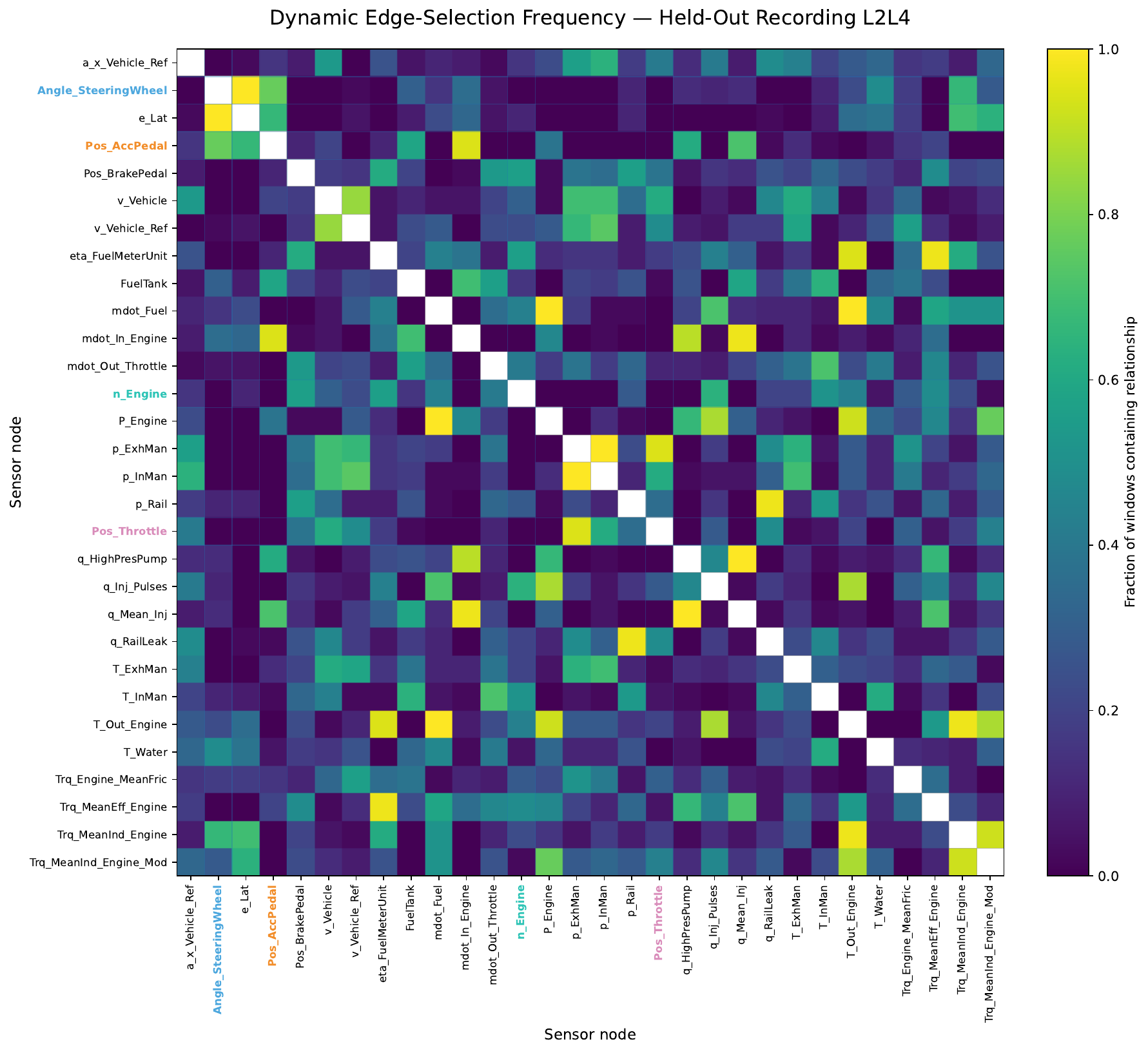}
    \caption{
        Edge-selection frequency across the 39 windows of the held-out
        $L2L4$ recording. High values correspond to persistent
        relationships, while intermediate values reflect
        state-dependent graph changes.
        \label{fig:dynamic_edge_frequency}
    }
\end{figure}

The heatmap in Figure~\ref{fig:dynamic_edge_frequency} reveals a sparse
but structured pattern. The relationship between
\texttt{Angle\_SteeringWheel} and \texttt{e\_Lat}, for example, is
selected frequently, while recurring connections also appear among
several engine- and fuel-related variables, including
\texttt{mdot\_Fuel}, \texttt{mdot\_In\_Engine},
\texttt{q\_HighPresPump}, and \texttt{q\_Mean\_Inj}. Other
relationships occur only intermittently. This mixture is consistent
with the intended behavior of the model: stable dependencies can be
retained, while less persistent neighborhoods change with the current
temporal state.

The active $L4$ node, \texttt{Pos\_Throttle}, provides a useful example.
Its neighborhood includes manifold-, vehicle-, and
temperature-related variables, but the same connections are not
maintained throughout the recording. A static graph would enforce one
neighborhood for all windows; the dynamic formulation does not.

\begin{figure}[H]
    \centering
    \includegraphics[width=\textwidth]
    {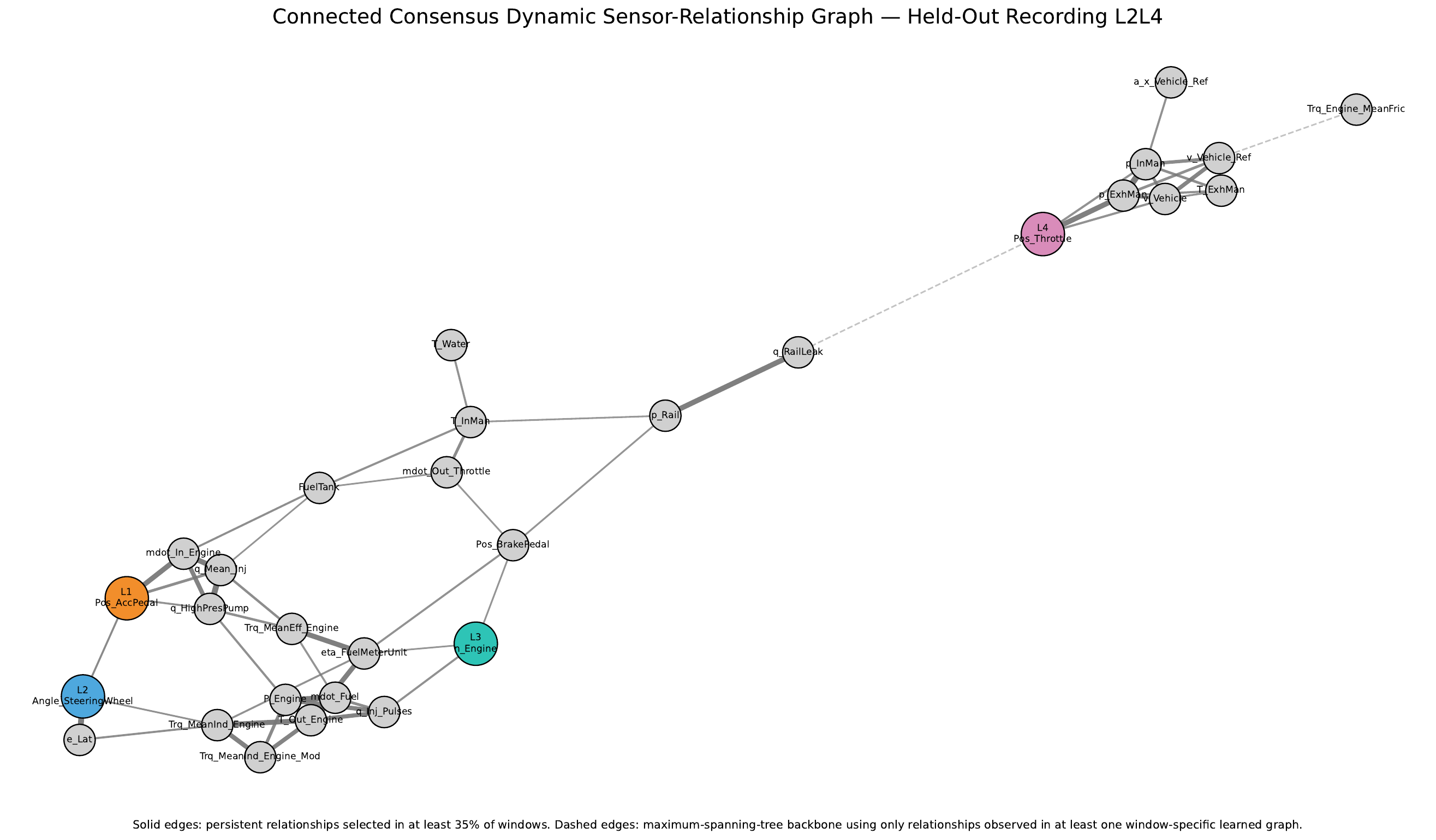}
    \caption{
        Consensus graph for the held-out $L2L4$ recording. Solid edges
        were selected in at least 35\% of the window-specific graphs.
        Dashed edges form an additional connectivity backbone using only
        relationships observed in at least one learned graph.
        \label{fig:consensus_dynamic_graph}
    }
\end{figure}

Figure~\ref{fig:consensus_dynamic_graph} summarizes the most persistent
part of this structure. The resulting graph highlights repeatedly
selected local groups while preserving weaker connections required to
visualize the overall topology. Together with the individual graphs and
frequency heatmap, it shows that the model learns a combination of
stable and transient sensor relationships rather than relying on a
single global structure. These visualizations should be interpreted as views of the predictive
representation learned for the RCA objective. They provide insight into
which sensor relationships support message passing, but they do not
establish physical or causal dependencies between the corresponding
variables.

To verify that the observed topology changes are not limited to the
illustrative $L2L4$ example, graph dynamicity was additionally
quantified across all nine held-out recordings. For consecutive
windows, the similarity between the corresponding undirected edge sets
was measured using the Jaccard coefficient,

\begin{equation}
J_t =
\frac{
|\mathcal{E}^{(t)} \cap \mathcal{E}^{(t+1)}|
}{
|\mathcal{E}^{(t)} \cup \mathcal{E}^{(t+1)}|
}.
\end{equation}

\begin{figure}[H]
    \centering
    \includegraphics[width=0.85\textwidth]
    {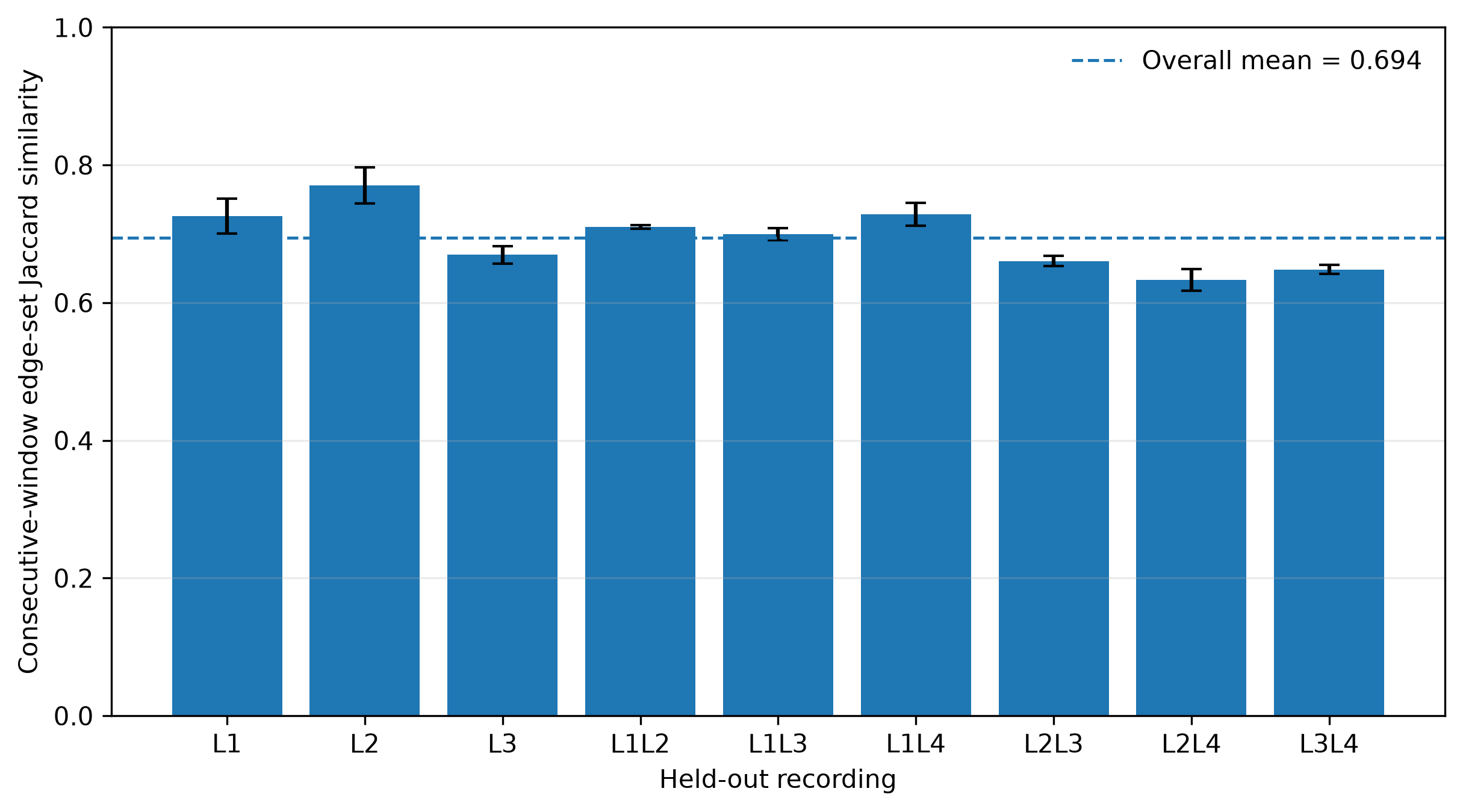}
    \caption{Mean consecutive-window edge-set Jaccard similarity for
    the nine held-out recordings. Error bars indicate variability across
    the three random seeds, and the dashed line denotes the overall mean.}
    \label{fig:graph_dynamicity}
\end{figure}

As shown in Figure~\ref{fig:graph_dynamicity}, the overall mean
consecutive-window Jaccard similarity is approximately $0.694$.
The learned graphs therefore retain a substantial fraction of their
relationships between neighboring windows while still undergoing
consistent topology changes. The degree of persistence also varies
across recordings, indicating that graph adaptation depends on the
observed system state rather than following a fixed transition pattern.
This quantitative analysis supports the intended behavior of the
proposed architecture: the learned sensor graph is dynamic, but its
evolution remains structured rather than arbitrary.

\subsection{Computational Efficiency and Practical Implications}
\label{subsec:complexity_results}

The performance gain of the dynamic formulation comes with only a moderate increase in model complexity. As summarized in Table~\ref{tab:complexity}, the proposed model contains approximately $40.2$k effective trainable parameters, compared with $35.9$k for the Static GCN and $36.2$k for the Static GAT.

Relative to the Static GCN, the proposed model uses only about $12\%$ more parameters while increasing recording-level Macro-F1 from $0.8911$ to $0.9524$. Single-window inference requires approximately $0.736\,\mathrm{ms}$, remaining below the $0.886\,\mathrm{ms}$ of the Static GAT despite the additional graph-construction step. Processing all 39 windows of one recording in a single batch takes approximately $0.832\,\mathrm{ms}$; this value reflects batch throughput rather than the response time of an individual window. From a practical perspective, dynamic graph construction therefore does not introduce a prohibitive computational burden. The additional cost
is small relative to the improvement in root-cause identification, and the model remains lightweight enough for offline HiL analysis and potentially for near-real-time diagnostic support.

\begin{table}[H]
\caption{Model complexity and inference latency measured on an NVIDIA
RTX PRO 6000 Blackwell Server Edition GPU. The 39-window value
represents amortized batch throughput and should not be interpreted as
single-window latency.
\label{tab:complexity}}

\begin{tabularx}{\textwidth}{lCCCC}
\toprule

\textbf{Model} &
\textbf{Parameters} &
\shortstack{\textbf{Batch-1}\\\textbf{[ms]}} &
\shortstack{\textbf{39 Windows}\\\textbf{[ms]}} &
\shortstack{\textbf{9-Fold Runtime}\\\textbf{[s]}} \\

\midrule

CNN--LSTM
& 104,900
& 0.218
& 0.221
& $10.3 \pm 2.0$ \\

Static GCN
& 35,905
& 0.653
& 0.727
& $20.6 \pm 0.9$ \\

Static GAT
& 36,161
& 0.886
& 0.922
& $21.0 \pm 1.6$ \\

\textbf{DySR-GNN}
& \textbf{40,193}
& \textbf{0.736}
& \textbf{0.832}
& $\boldsymbol{21.2 \pm 1.7}$ \\

Context-aware variant
& 52,673
& 0.758
& 0.843
& $20.4 \pm 1.0$ \\

Gated context-aware variant
& 56,834
& 0.850
& 0.939
& $21.1 \pm 0.9$ \\

\bottomrule
\end{tabularx}
\end{table}

\section{Conclusions}
\label{sec:conclusion}

This work introduced a Dynamic Sensor-Relationship GNN (DySR-GNN) for supervised
multi-label root-cause analysis in automotive HiL time series. The model
combines sensor-wise temporal encoding with trainable sensor identities
and reconstructs a sparse graph for each input window before applying
graph message passing. Across the recording-disjoint evaluation, the
proposed model achieved the best overall performance, reaching a
recording-level Macro-F1 of $0.9524$, an MCC of $0.9302$, and an
exact-match accuracy of $0.8519$. The comparison with CNN--LSTM, Static
GCN, and Static GAT baselines shows that adaptive relational structure
provides useful information beyond temporal modeling or a fixed graph
topology.

The learned graphs further reveal a mixture of persistent and
state-dependent sensor relationships. These structures should be
interpreted as task-relevant predictive relationships rather than causal
dependencies. A current limitation is that the RCA formulation assumes a
predefined set of candidate root locations and uses windows extracted
from the known fault-active interval. Open-set root causes, fault-onset
identification, and transfer beyond the controlled HiL environment are
therefore not addressed in the present study.

Future work should investigate these settings while extending the
evaluation to additional fault locations, operating conditions, and
automotive platforms. Alternative graph-construction strategies and the
integration of causal knowledge also provide
promising directions for developing more generalizable RCA models.

\vspace{6pt} 





\authorcontributions{
Conceptualization, H.O.; methodology, H.O.; software, H.O.; validation, H.O.; 
formal analysis, H.O.; investigation, H.O.; resources, H.O. and M.A.; 
data curation, H.O.; writing---original draft preparation, H.O.; 
writing---review and editing, H.O. and M.A.; visualization, H.O.; 
supervision, M.A. and A.R.; project administration, H.O., M.A. and A.R. 
All authors have read and agreed to the published version of the manuscript.
}

\funding{We acknowledge support by the Open Access Publishing Fund of Clausthal University of Technology.}

\institutionalreview{Not applicable.}

\informedconsent{Not applicable.}

\dataavailability{Data are available on request due to restrictions.}

\begin{adjustwidth}{-\extralength}{0cm}

\reftitle{References}




\bibliography{references}
%


\PublishersNote{}
\end{adjustwidth}
\end{document}